\documentclass[conference]{IEEEtran}
\IEEEoverridecommandlockouts
\usepackage{cite}
\usepackage{amsmath,amssymb,amsfonts}
\usepackage{bm}
\usepackage{bbm}
\usepackage{algorithmic}
\usepackage{booktabs}
\usepackage{graphicx}
\usepackage{textcomp}
\usepackage{soul}
\usepackage{xcolor}
\usepackage{stfloats}
\usepackage{balance}
\usepackage{subcaption}

\newcommand\x{\bm{x}}

\newcommand\lambdab{\boldsymbol{\lambda}}
\newcommand\A{\bm{A}}

\newcommand\X{\bm{X}}

\newcommand\Xt{\bm{\mathcal{X}}}

\newcommand\RR{\mathbb{R}}

\definecolor{JungleGreen}{rgb}{0,0.7,0.4}
\definecolor{deepPurple}{rgb}{0.5,0.1,0.5}

\renewcommand\min[1]{\underset{#1}{\mathrm{min}}\,}
\renewcommand\max[1]{\underset{#1}{\mathrm{max}}\,}

\def\BibTeX{{\rm B\kern-.05em{\sc i\kern-.025em b}\kern-.08em
    T\kern-.1667em\lower.7ex\hbox{E}\kern-.125emX}}
\begin{document}

% Efficient Nonparametric PDF Estimation via MDL and Tensor Factorization

% Quantile-Based MDL and Tensor Factorization for Density Estimation

% Adaptive Binning and Tensor Decomposition for Improved PDF Estimation
\title{A Unified MDL-based Binning and Tensor Factorization Framework for PDF Estimation\\
\thanks{The authors gratefully acknowledge the support of the German Research Foundation (DFG) under the PROMETHEUS project (reference no. HA 2239/16-1, project no. 462458843).}
}

\iffalse
\author{\IEEEauthorblockN{Mustafa Musab}
\IEEEauthorblockA{\textit{Communications Research Laboratory} \\
\textit{Ilmenau University of Technology}\\
Ilmenau, Germany \\
mustafa.musab@tu-ilmenau.de}
\and
\IEEEauthorblockN{Joseph K. Chege}
\IEEEauthorblockA{\textit{Communications Research Laboratory} \\
\textit{Ilmenau University of Technology}\\
Ilmenau, Germany \\
joseph.chege@tu-ilmenau.de}
\and
\IEEEauthorblockN{Arie Yeredor}
\IEEEauthorblockA{\textit{School of Electrical Engineering} \\
\textit{Tel Aviv University}\\
Tel Aviv, Israel \\
ariey@tauex.tau.ac.il}
\and
\IEEEauthorblockN{Martin Haardt}
\IEEEauthorblockA{\textit{Communications Research Laboratory} \\
\textit{Ilmenau University of Technology}\\
Ilmenau, Germany \\
martin.haardt@tu-ilmenau.de}
}
\fi

\author{
	\IEEEauthorblockN{\emph{Mustafa Musab$^1$, Joseph K.~Chege$^1$, Arie Yeredor$^2$, and Martin Haardt$^1$}
	}
	\IEEEauthorblockA{$^1$Communications Research Laboratory, Ilmenau University of Technology, Ilmenau, Germany
	}
	\IEEEauthorblockA{$^2$School of Electrical Engineering, Tel Aviv University, Tel Aviv, Israel
	}
    \IEEEauthorblockA{Email: \{mustafa.musab, joseph.chege, martin.haardt\}@tu-ilmenau.de, ariey@tauex.tau.ac.il
	}
}

\maketitle

\begin{abstract}
Reliable density estimation is fundamental for numerous applications in statistics and machine learning. In many practical scenarios, data are best modeled as mixtures of component densities that capture complex and multimodal patterns. However, conventional density estimators based on uniform histograms often fail to capture local variations, especially when the underlying distribution is highly nonuniform. Furthermore, the inherent discontinuity of histograms poses challenges for tasks requiring smooth derivatives, such as gradient-based optimization, clustering, and nonparametric discriminant analysis.
In this work, we present a novel non-parametric approach for multivariate probability density function (PDF) estimation that utilizes  minimum description length (MDL)-based binning with quantile cuts. Our approach builds upon tensor factorization techniques, leveraging the canonical polyadic decomposition (CPD) of a joint probability tensor. We demonstrate the effectiveness of our method on synthetic data and a challenging real dry bean classification dataset.
\end{abstract}


\section{Introduction}
Accurate density estimation is crucial in many areas of statistics and machine learning, including clustering, classification, and signal processing.  In particular, mixture models, where the overall distribution is expressed as a weighted sum of component densities, play a key role in modeling complex, multimodal data. 
% Recovering both the component ..
Recovering both the component densities and their corresponding weights is essential for revealing the underlying distribution and making reliable inferences.
% Many natural processes are inherently ..

Although many natural processes are inherently continuous, conventional methods, such as uniform histograms or probability mass functions (PMFs), often fail to capture local variations in the data.
This is especially true in applications involving complex and multimodal distributions. Examples include biological measurements (e.g., gene expression levels in cancer genomics, and multimodal intensity distributions in medical imaging like the BrainWeb MRI dataset \cite{Cocosco1997BrainWebOI}) and agricultural applications. 
% For instance, in agricultural contexts,  the classification of dry beans relies on physical characteristics such as appearance, size, and internal quality—each exhibiting multimodal distributions~\cite{koklu2020multiclass}. However,
% manual classification is labor-intensive and prone to errors,
% motivating the need for methods that can reliably capture
% complex, multimodal distributions.

% Tensor-based density estimation
Recent advances in tensor-based density estimation have attracted considerable attention in the statistical community (e.g., Anandkumar~\textit{et al.}~\cite{anandkumar2014tensor}; Miranda~\textit{et al.}~\cite{miranda2018tprm}; Gottesman~\textit{et al.}~\cite{gottesman_weighted_nodate}). In all these works, the focus is on fully parametric models in which the mixture components are assumed to belong to a specific parametric family. In contrast, our method makes no parametric assumption about the underlying density. 
% In contrast, our study does not assume any parametric form for the underlying distribution.
% Our approach is most similar to that of Kargas

Our approach is most similar to that of Kargas~\textit{et al.}~\cite{kargas2019learning}, who proposed a tensor–based method for learning mixtures. In their work, the dataset is discretized using uniform bins, followed by tensor factorization to recover the discretized PDFs, and finally, sinc interpolation is used to reconstruct the continuous PDF. However, uniform bins have been shown to be effective only when the data are approximately uniformly distributed~\cite{rissanen1992density}. In contrast, by employing minimum description length (MDL)-based binning with quantile cuts, our method overcomes these limitations. 
% the minimum description length (MDL) framework

Much of the existing literature within the MDL framework has focused on the estimation of histogram density, producing discrete PMF models (e.g.,~\cite{kontkanen2007mdl}). Although such methods yield an accurate discrete representation, they do not directly address PDF estimation. In this work, we extend the classical MDL framework to continuous multivariate settings.
Moreover, while previous work~\cite{amiridi2022low} represents the joint PDF using a low-rank tensor in the Fourier domain, our method operates directly in the data domain.
% Evaluating the quality of density models is an open and
% difficult problem [45]. Following the approach in [15], [35], we calculate and report the average log-likelihood of unseen data samples (testing set), further averaged over 5 random data splits.

\iffalse
\subsection{Notation}
We denote a scalar, a vector, a matrix and a tensor by $x$, $X$, $\X$, and $\Xt$, respectively. The outer product between vectors is denoted by $\circ$, and the transpose of a vector or matrix is indicated by the superscript $^\mathsf{T}$. PMF and PDF are represented by $p(\cdot)$ and $f(\cdot)$, respectively. Additional notation specific to our MDL and tensor factorization framework is introduced as needed in subsequent sections.
\fi
\section{Problem Formulation} \label{sect2}
Consider a collection of $N$ continuous random variables $\mathsf{X} = \{X_1, \dots, X_N\}$. 
Furthermore, given a realization $\x=\{x_1,\dotsc,x_N\}$ of $\mathsf{X},$ assume that the joint PDF $f_{\mathsf{X}}(\x)$ can be written as a weighted sum of $R$ \footnote{The choice of the tensor rank $R$ will be discussed in Subsection \ref{subsetc: real data}.} conditional PDFs $f_{\mathsf{X}\,|\,H}$, and that each conditional PDF can be factorized into a product of its marginal densities such that
\begin{equation} \label{eq:jointPDF}
   f_{\mathsf{X}}(x_1,\dotsc,x_N) = \sum_{r=1}^R p_H(r) \prod_{n=1}^N f_{X_n|\,H}(x_n\,|\,r).
\end{equation}
The expression in \eqref{eq:jointPDF} represents $f_{\mathsf{X}}(\x)$ as a mixture of product distributions, where $H$ can be interpreted as a latent variable taking $R$ states, while $p_H(r)$ is the prior probability of selecting the $r$-th product in the mixture (e.g., \cite{kargas2019learning}).
% For an \(N\)-dimensional tensor with full-rank factor matrices 
% \(\mathbf A_n\in\mathbb R^{I_n\times R}\), \(\forall n\), the Kruskal condition 
% \(\sum_n \min(I_n,R)\ge2R+(N-1)\) is sufficient for a unique decomposition\footnote{In our present case the full-rank condition holds and hence the maximum possible rank is \(R=23\).  We will explain our choice of \(R\) in Section~\ref{sec:rank-selection}.}.

% \revOne{The choice of the tensor rank $R$ will be determined empirically in Sec. \ref{sub. real data} via variational Bayesian inference (VB-PMF), which automatically selects an effective $R$.}
%
However, note that no explicit parametric form (e.g., Gaussian, etc.) is specified for the conditional PDFs.

%

%We adaptively discretize each $X_n$ into $I_n$ bins using the Minimum Description Length (MDL) criterion. 
%
%As a result, each $X_n$ is mapped to a discrete label $\widetilde{X}_n \in \{1,\dots, I_n\}$ where $I_n$ is the estimated number of bins for variable $n$. The resulting joint PMF of $X$ can be organized as an $N$-way tensor $\bm{\mathcal{X}} \in \mathbb{R}^{I_1 \times \dots \times I_N}$, where 
%\[\bm{\mathcal{X}}(i_1, \dotsc, i_N) = \mathsf{Pr}(\widetilde{X}_1 = i_1, \dotsc, \widetilde{X}_N = i_N)\]
The support of each $X_n$ is discretized into $I_n$ bins $\Delta_n^{i_n}$, $i_n = 1, \dotsc, I_n$, resulting in a discretized version of \eqref{eq:jointPDF} which can be conveniently represented by an $N$-dimensional tensor $\Xt \in \mathbb{R}^{I_1 \times \dots \times I_N}$ where
\begin{equation}
    \begin{aligned}[b]
        \Xt(i_1,\dotsc,i_N) &= \mathsf{Pr}(X_1 \in \Delta_1^{i_1}, \dotsc, X_N \in \Delta_N^{i_N}) \\ 
        & = \sum_{r=1}^R p_H(r) \prod_{n=1}^N \mathsf{Pr}(X_n \in \Delta_{n}^{i_n}\,|\,H=r).
    \end{aligned}    
\end{equation}
%We further assume that $\bm{\mathcal{X}}$ admits a low-rank canonical polyadic decomposition (CPD) of rank $R$ \cite{sidiropoulos_tensor_2017}:
By defining $\A_n(i_n,r) = \mathsf{Pr}(X_n \in \Delta_{n}^{i_n}\,|\,H=r)$ and $\lambda_r = p_H(r)$, it can be observed that $\Xt$ admits a rank-$R$ canonical polyadic decomposition (CPD) \cite{kolda2009tensor, sidiropoulos2017tensor} with factor matrices $\A_n \in \RR^{I_n \times R}$ and a ``loading vector" $\lambdab \in \RR^R$, subject to a set of probability simplex constraints, i.e.,
\begin{equation} \label{eq:CPD}
    \begin{aligned}
    \Xt  &=  \sum_{r=1}^R \lambda_r \A_1(:,r) \circ \A_2(:,r) \circ \cdots \circ \A_N(:,r) \\
    & \mathrm{subject~ to~~} \bm{\lambda} > \bm{0}, ~ \bm{1} ^ \mathsf{T} \bm{\lambda} = 1 \\
   & \qquad \qquad \quad \A_n \ge \bm{0}, ~ \bm{1} ^ \mathsf{T} \A_n = \bm{1} ^ \mathsf{T}, ~ n = 1, \dotsc, N,
    \end{aligned}
\end{equation}
where $\circ$, $(\cdot)^\mathsf{T}$, and $\bm{1}$ represent the outer product, the transpose operator, and an all-ones vector, respectively.
It has been shown in \cite{kargas2018tensors} that \eqref{eq:CPD} corresponds to a na\"{i}ve Bayes model with a root variable $H$ that takes a finite number $R$ of states.

%Given this partially observed data,
Given a dataset of $T$ independent and identically distributed realizations $\x_t=\{x_{1,t},\dotsc,x_{N,t}\}$ ($t=1,\dotsc,T$) of $\mathsf{X}$, our objective is to recover the underlying continuous PDF $f_{\mathsf{X}}(\x)$ from an estimate of the discretized PDF (joint PMF) $\Xt$.
Leveraging on the na\"{i}ve Bayes structure of $\Xt$, we propose to recover $f_{\mathsf{X}}(\x)$ by interpolating each discretized marginal distribution $\A_n(:, r) = p(X_n\,|\,H=r)$ separately, followed by recombination of the interpolated marginal PDFs to form the joint PDF.
A similar approach was considered in \cite{kargas2019learning}, where discretization was carried out using uniform bins, while sinc interpolation was adopted to recover the marginal PDFs.
However, in the following section, we propose a PDF estimation method that employs nonuniform bins whose width and number are selected to minimize an MDL criterion.
The resulting nonuniform bins necessitate the use of a different interpolation strategy.
%
%We provide motivating examples illustrating the advantages of our approach compared to uniform binning.
We present motivating examples highlighting the advantages of our approach over uniform binning.

\section{Proposed Methodology} \label{sect3}
We propose a PDF estimation framework that addresses the limitations of uniform binning and sinc interpolation. Our approach proceeds in three main steps. We first employ an MDL-based strategy to learn the histogram of each marginal distribution. By determining both the number and locations of the bin edges in a data-driven manner, this step captures the inherent structure of the data. As a result, continuous variables are effectively transformed into categorical ones. Next, we recover the complete discretized PDF by applying a maximum likelihood PMF estimation algorithm (SQUAREM-PMF, \cite{chege2022efficient}) within a coupled nonnegative tensor factorization framework. Finally, we employ spline interpolation to obtain a smooth PDF from the discretized joint PDF estimate. Although this description emphasizes PDF estimation, the same adaptive binning and smoothing procedure naturally extends to nonparametric mixture models.
% In particular, the MDL-based histogram can capture multiple modes or subpopulations within the data without assuming a fixed parametric form.

\subsection{MDL binning with quantile cuts}
Many histogram density estimation methods rely on uniform binning, which can be suboptimal if the underlying distribution is strongly nonuniform or multimodal. Intuitively speaking, wider bins in regions with sparse data help to reduce noise from sampling randomness, whereas narrower bins in dense regions capture fine details more effectively. Therefore, adapting the bin widths and locations to the data can significantly improve estimation quality.
In the MDL framework~\cite{kontkanen2007mdl}, the goal is to select the simplest possible model that sufficiently explains the observed data by determining both the optimal number of bins and their locations. This dual optimization is formalized via the \emph{normalized maximum likelihood} (NML), which provides strong theoretical guarantees~\footnote{The NML criterion provides two important theoretical guarantees:  
(i) it uniquely solves Shtarkov’s minimax problem~\cite{shtar1987universal}, (ii) it also minimizes the expected worst-case regret in code length among all universal codes\cite{rissanen2002strong}.}. 

Let $\mathcal{M}= \bigl\{\,f(\,\cdot\,\mid\theta):\theta\in\Theta\bigr\}$ denote a histogram model class (for example, all histograms with $K$ bins). Each $\theta \in \Theta$ represents a specific choice of bin edges, thereby defining a particular histogram within this class. We use  $f\bigl(\x\mid\theta\bigr)$ to denote the density given specific parameters, and $f\bigl(\x\mid\theta,\mathcal{M}\bigr)$ explicitly highlights the density given these parameters within the chosen histogram model class $\mathcal{M}$.
For a univariate sample $\x= \{x_{1},\dots,x_{T}\}\subset X$
of length $T$, the maximum-likelihood estimate is $\hat{\theta}(\x)
\;=\;
\arg\max{\theta\in\Theta}
f\bigl(\x\mid\theta\bigr)$.

The NML density
\cite{shtar1987universal} is defined by
\begin{equation}
    f_{\text{NML}}\!\bigl(\x \mid\mathcal{M}\bigr)
    =
    \frac{f\!\bigl(\x \mid\hat{\theta}(\x),\mathcal{M}\bigr)}
     {\mathcal{R}_{\mathcal{M}}}
    \label{eq:NML},
\end{equation}
where the normalizing constant
$\mathcal{R}_{\mathcal{M}}$, known as the \emph{parametric complexity}, quantifies the intrinsic complexity of the model:
% \begin{equation}
% R_{n}(\mathcal{M})
% =\int_{X^{T}}
%      f\!\bigl(\mathbf y^{n}\mid\hat{\theta}(\mathbf y^{n}),\mathcal{M}\bigr)
%      \,\mathrm d\mathbf y^{n},
% \end{equation}
\begin{equation} \mathcal{R}_{\mathcal{M}} = \int_{\x \in X} f(\x\mid \hat{\theta}(\x), \mathcal{M}) \, \mathrm{d}\x.
\end{equation}
The \emph{stochastic complexity} of $\x$ under $\mathcal{M}$ is then given by
\[
\mathrm{SC}\bigl(\x \mid\mathcal{M}\bigr)
=
-\log f_{\text{NML}}\!\bigl(\x \mid\mathcal{M}\bigr),
\label{eq:SC}
\]
and the MDL principle prescribes selecting model $\mathcal{M}^{\star}$ and the corresponding $\hat{\theta}(\x)$ that minimizes this quantity\cite{rissanen1986stochastic}.
% toy example: distribution
% toy example: distribution
% toy example: distribution
\begin{figure}[t]
\centering
\centerline{\includegraphics[width=0.48\textwidth]{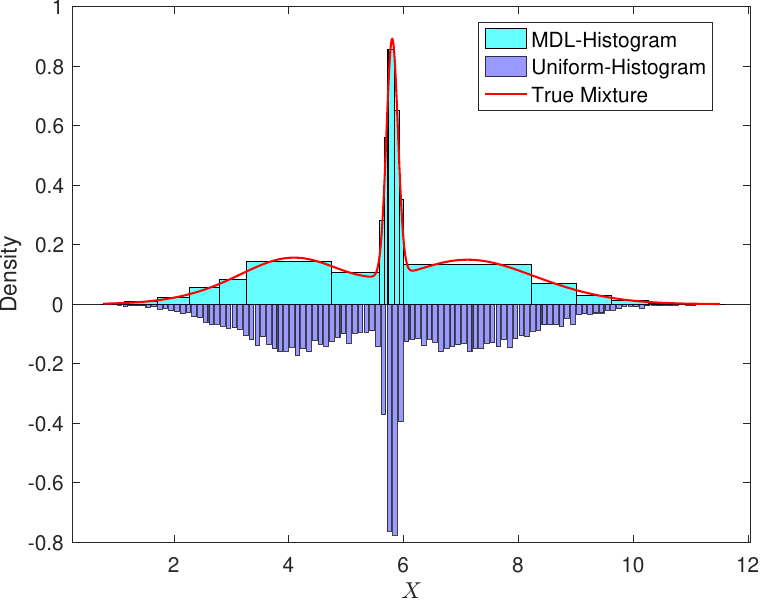}}
\caption{The generating density, the MDL-optimal histogram (19 bins), and the uniform 100-bin histogram (mirrored).}
\label{fig:mdl_emp}
\end{figure}

% MDL cuts
% MDL cuts
An essential step in constructing the MDL-optimal histogram is the definition of candidate cut points for the optimization process. The authors in \cite{kontkanen2007mdl} describe two approaches for selecting candidate cuts. The first, known as the midpoint approach, places a cut at the midpoint between each pair of consecutive data points. While this method ensures that every bin has at least one observation, it does not allow for empty bins, which is a disadvantage when large gaps are present. An alternative approach involves placing two cut points between each pair of consecutive data values, positioned as close as possible to the data values. Although this method improves adaptability to data gaps, it significantly enlarges the candidate-cut set, increasing the computational cost. 
% quantile cuts

Unlike \cite{kontkanen2007mdl}, we adopt a \emph{quantile}-based strategy: candidate cuts are the empirical quantiles, giving equal-frequency bins that mirror the underlying density.  This data-adaptive placement captures structure more faithfully while remaining computationally efficient.
% Unlike \cite{kontkanen2007mdl}, we adopt a \emph{quantile} strategy, leveraging the fact that equal-frequency partitions naturally reflect the underlying density \cite{sulewski2021equal}. Candidate cuts  are chosen based on empirical quantiles, ensuring an approximately equal distribution of data points within each bin. This data-adaptive placement captures structure more faithfully while remaining computationally efficient.\\
Let $x_{(1)} \;\le\; x_{(2)} \;\le\; \cdots \;\le\; x_{(T)}$ denote an ordered sample of size $T$. The empirical cumulative distribution function (eCDF)  is defined as:
\[
\hat{F}_{T}(y) 
\;=\; 
\frac{1}{T}\sum_{t=1}^{T} \mathbbm{1}\{x_{(t)} \le y\},
\]
where $\mathbbm{1}(\cdot)$ is the indicator function. For a given probability \( p \in [0,1] \), the empirical quantile $Q(p)$ is defined as the smallest $y$ such that $\hat{F}_t(y)\ge p$. In our application, to partition the support of the data into $E$ equal-frequency bins, we define the set of candidate interior cuts as:
% \[
% \tilde C \;=\;
% \Bigl\{\,c_{j}=Q(j/E)=\hat F_{T}^{-1}(j/E)\Bigr\}_{j=1}^{E-1},
% \]
\[
    \tilde{\mathcal{C}} = \left\{ c_j = Q\left( \frac{j}{E} \right) = \hat{F}_T^{-1}\left( \frac{j}{E} \right) \right\}_{j=1}^{E-1}
\]

The goal is then to select a \(K\)-bin subset \(C\subseteq\tilde C\) that minimizes the MDL criterion:
\begin{equation} 
    \label{eq:mdl score}
    B(\x \mid E,K,C)
    = \mathrm{SC}(\x \mid C)\;+\;\log\binom{E}{K-1},
\end{equation}
where $\mathrm{SC}(\x \mid C)$  is the stochastic complexity (negative log NML), measuring how well the model fits the observed data. The second term, $\log\binom{E}{K-1}$, represents the model complexity penalty. It measures the description length needed to specify which $K-1$ cut points are chosen from the $E$ possible candidates.
% The second term represents the codelength for encoding the cuts, which is
% described as the number of ways to choose the cut points.
As detailed in~\cite{kontkanen2007mdl}, the $\mathrm{SC}$ is computed recursively, and the optimal cuts can be found by dynamic programming in $\mathcal{O}(E^2 \cdot K_{\text{max}})$ time, where $K_{\text{max}}$ is the maximum number of bins considered during optimization.

\begin{figure}[t]
\centering
\centerline{\includegraphics[width=0.48\textwidth]{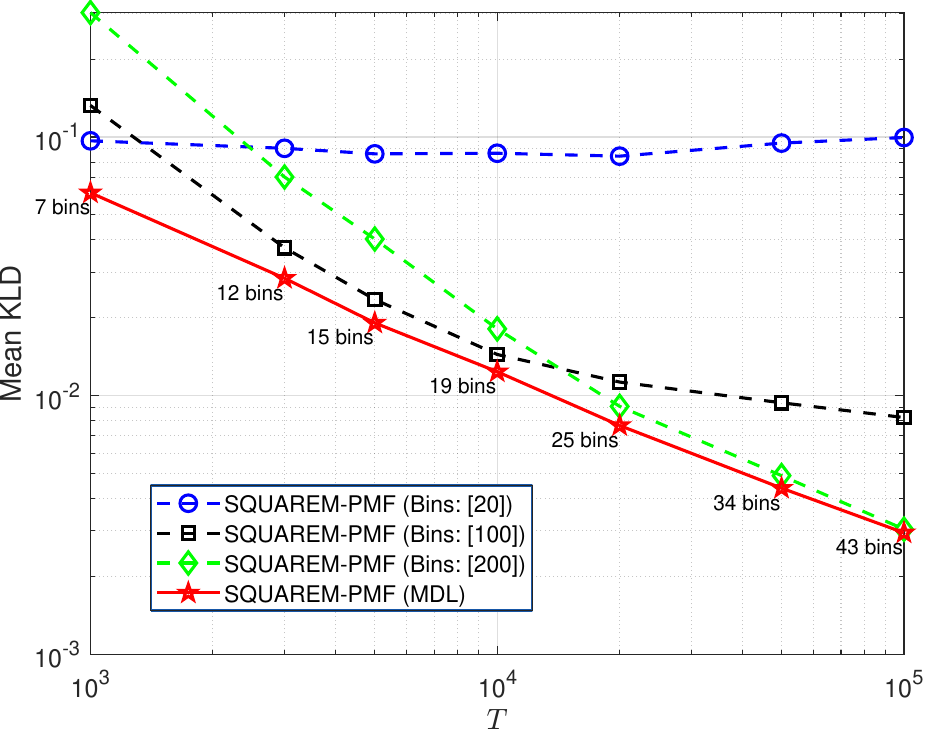}}
\caption{KLD between the true density and its estimates obtained with the MDL histogram and uniform histograms of 20, 100, and 200 bins.}
\label{fig:kld_mdl_pm}
\end{figure}

% To demonstrate the advantage
To demonstrate the advantage of the MDL histogram, we consider a toy example of five-component univariate Gaussian mixture. In our experiment, samples are drawn from the mixture depicted in Fig.~\ref{fig:mdl_emp}. 
% Fig.~\ref{fig:kld_mdl_pm} presents the Kullback–Leibler
Fig.~\ref{fig:kld_mdl_pm} shows the mean Kullback–Leibler divergence (KLD) (over 50 trials) between the true and estimated densities.
Despite using significantly fewer bins, the MDL histogram consistently achieves a lower KLD compared to its uniform counterparts. 
% As the sample size increases
As the sample size grows, the performance of the 100 and 200-bin uniform histograms gradually converges to that of the MDL. 
% Interestingly, at a sample size of 10,000
Interestingly, at a sample size of $10^4$, the 19-bin MDL histogram significantly outperforms the uniform 20-bin, despite the nearly identical bin count. This highlights that the placement of the bins is as important as the bin count in capturing the underlying distribution.

% toy example kls

\subsection{Estimation of the Discretized PDF}
We estimate the joint PMF (discretized PDF) by applying a low-rank factorization to the discretized data. Specifically, we adopt the SQUAREM-PMF algorithm proposed in \cite{chege2022efficient}, which extends the expectation-maximization (EM) algorithm proposed in \cite{yeredor2019maximum} with a squared iterative methods (SQUAREM) acceleration step, thereby improving convergence speed even when the data are only partially observed. In this framework, the joint PMF tensor $\bm{\mathcal{X}}$ is constrained to have a CPD of rank $R$. The discretized observations are used in a maximum-likelihood (ML) setting \cite{yeredor2019maximum}:
\begin{equation} \label{eq:ml_obj} 
    \begin{aligned}
    \min{\{\bm{A}_n\}_{n=1}^{N}, \bm{\lambda}} \quad & - \sum_{t=1}^{T} \log \left( \sum_{r=1}^{R} \lambda_r \prod_{n=1}^{N} \bm{A}_n(x_{n,t}, r) \right) \\
    \mathrm{subject~ to~~} & \bm{\lambda} > \bm{0}, \quad \bm{1}^\mathsf{T} \bm{\lambda} = 1 \\
    & \bm{A}_n \geq \bm{0}, \quad \bm{1}^\mathsf{T} \bm{A}_n = \bm{1}^\mathsf{T}, \quad n = 1, \dots, N
    \end{aligned}
\end{equation}

\begin{figure*}[t]
    \centering
    \begin{minipage}{0.32\textwidth}
        \centering
        \includegraphics[width=\textwidth]{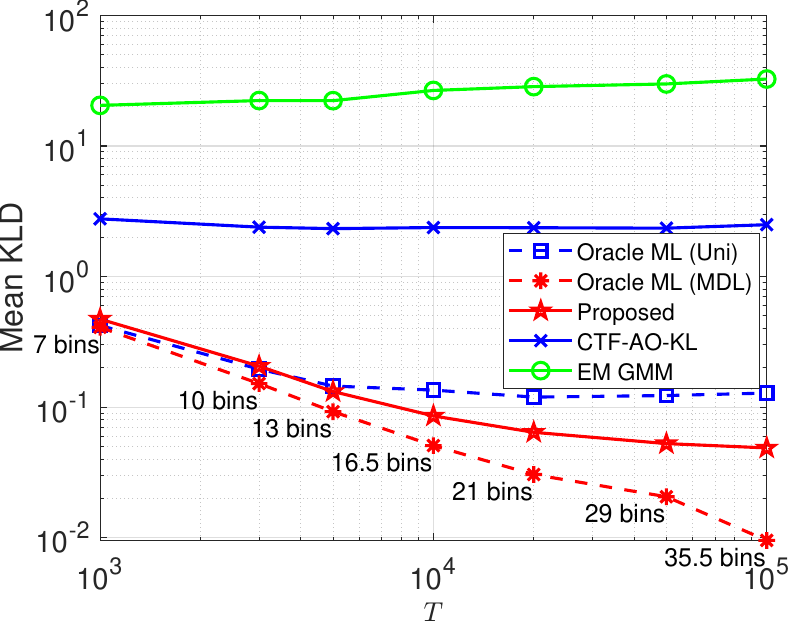}
    \end{minipage}
    \hfill
    \begin{minipage}{0.32\textwidth}
        \centering
        \includegraphics[width=\textwidth]{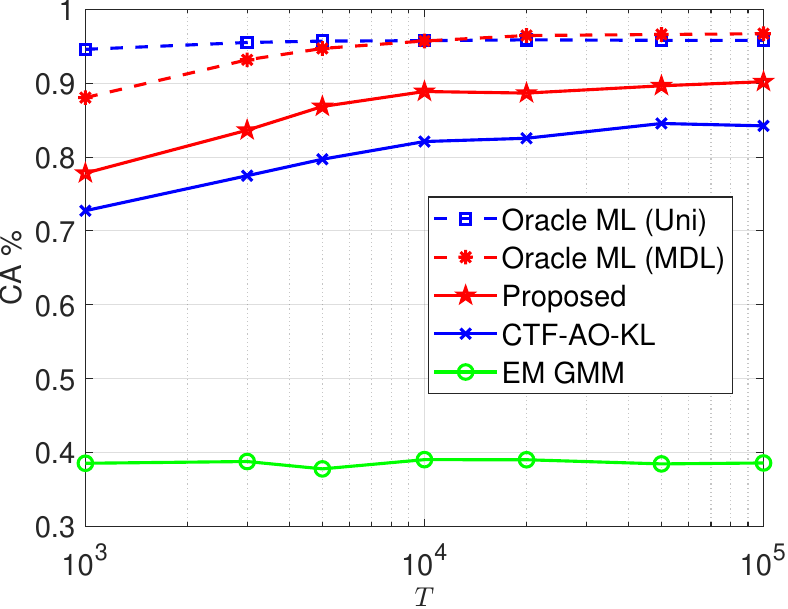}
    \end{minipage}
    \hfill
    \begin{minipage}{0.32\textwidth}
        \centering
        \includegraphics[width=\textwidth]{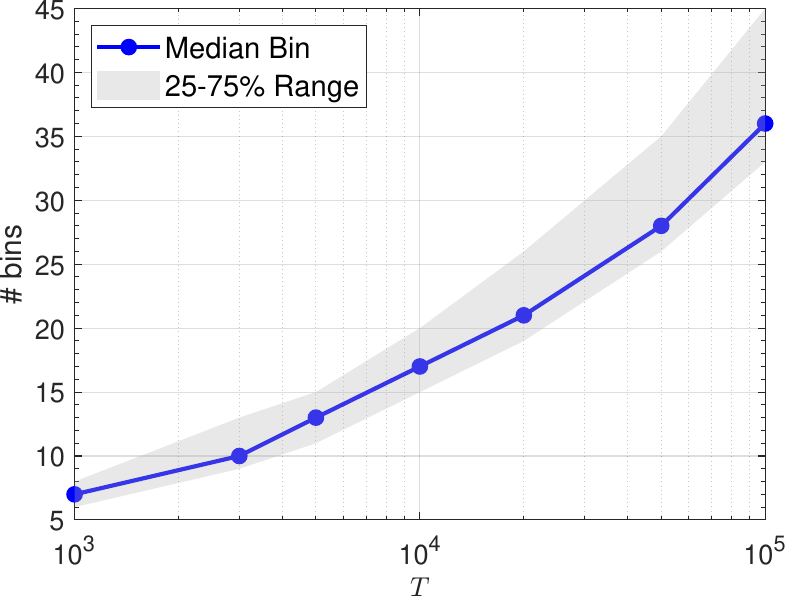}
    \end{minipage}

    \caption{Performance versus sample size: (a) KLD, (b) clustering accuracy (CA), and (c) median number of MDL-selected bins.}
    \label{fig:comparison}
\end{figure*}

The EM updates for $\{\bm{A}_n\}$ and $\lambdab$ are derived from the objective \eqref{eq:ml_obj}. However, EM typically exhibits slow convergence.

SQUAREM-PMF addresses this limitation by squaring the standard EM fixed-point updates. At each iteration, it computes two successive EM steps, then performs a polynomial extrapolation in parameter space to accelerate convergence--while preserving the monotonic likelihood increase guaranteed by EM. Numerical evidence in~\cite{chege2022efficient} shows that SQUAREM-PMF can substantially reduce the iteration count and run time compared to both plain EM and other factorization-based approaches, making it an efficient choice for the PMF estimation step in our proposed methodology.

\subsection{PDF Reconstruction using Spline Interpolation}
In the classical Shannon-sampling framework, a bandlimited signal can be perfectly reconstructed from uniform samples via sinc interpolation. In fact, Kargas \textit{et al.} \cite{kargas2019learning} demonstrated that a PDF which is (approximately) bandlimited with a cutoff frequency $\omega_c$  can be reconstructed from uniformly spaced samples of its corresponding CDF, provided the sampling interval $\Delta t \le\frac{\pi}{\omega_{c}}$. However, extending this result to our settings is problematic because the MDL-estimated bins are nonuniform; consequently, the underlying assumption of uniform sampling (and hence a time-invariant reconstruction kernel) does not hold. Although generalized non-uniform sampling theory provides conditions under which sinc interpolation can be applied to nonuniform samples (see, e.g.,~\cite{maymon2011sinc}), these require that the average sampling rate satisfies the Nyquist criterion, something we cannot ensure a-priori with MDL-chosen cuts. This limitation motivates the use of cubic spline interpolation, which naturally accommodates nonuniform spacing and produces smooth, continuous curves with continuous first and second derivatives (i.e., the $C^2$ class) \cite{deboor}. Let the $r$-th column $\A_n(:,r) \in \RR^{I_n}$ of the $n$-th factor matrix $\A_n$ represent the discretized conditional PDF $p_{X_n\,|\,H}(x_n\,|\,r)$ of the $n$-th random variable. The corresponding CDF points are obtained via the cumulative sum of that column, yielding $\{F_i\}_{i=1}^{I_n}$. For each bin $\Delta^i = [i, i+1]$, a cubic polynomial is fitted:
\begin{equation}
    S_i(x) = a_i + b_i(x - i) + c_i(x - i)^2 + d_i(x - i)^3.
\end{equation} This yields a piecewise definition of the continuous CDF
\begin{equation}
F_{X \mid H}(x_n \mid r) = \sum_{i=1}^{I_n-1} S_i(x_n) \cdot \mathbbm{1}_{[i, i+1)}(x_n),
% \tag{9}
\end{equation}
where
\[
\mathbbm{1}_{[i, i+1)}(x_n) =
\begin{cases}
    1, & \text{if } x_n \in [i, i+1), \\
    0, & \text{otherwise}.
\end{cases}
\]
% zero end-slope conditions
We enforce zero end-slope conditions (i.e., the first derivative of the spline is zero at the boundaries). This ensures that the reconstructed PDF smoothly approaches zero at the boundaries and prevents unrealistic extrapolation at the edges.

% The final PDF
The final PDF is obtained by differentiating the piecewise CDF approximation:
\[
\hat{f}_{X_n \mid H} (x_n \mid r) = \frac{\rm d}{\mathrm{d}x_n} F_{X_n \mid H} (x_n \mid r) 
\]

\section{Simulation Results} 
\subsection{Synthetic Data}
\begin{figure}[t]
    \centering    \centerline{\includegraphics[width=0.48\textwidth]{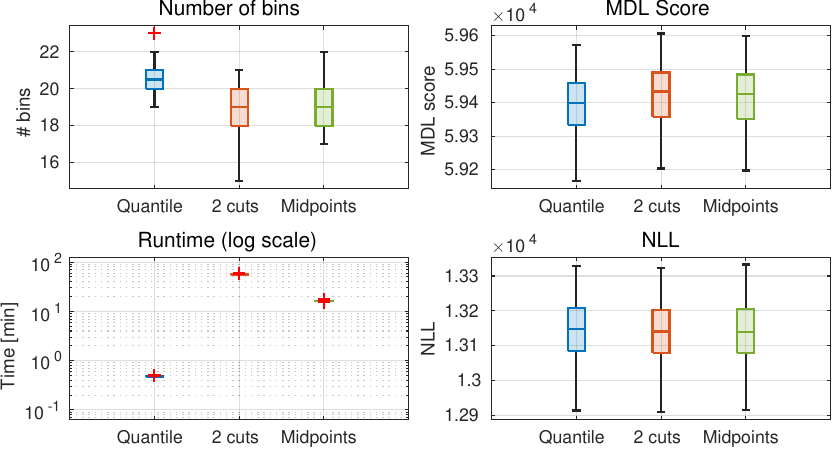}}
    \caption{Performance of Cuts Placement Strategies.}
    \label{fig:binning_strat}
\end{figure}

We first investigate how the candidate-cutting strategy affects the MDL algorithm. 
A six-component univariate Gaussian mixture is sampled with
$T=20\,000$ observations, the experiment is repeated 50 times.
% A univariate Gaussian mixture with $R=6$ components is sampled with $T=20\,000$ observations, and all results are averaged over 50 independent trials. 
Each method starts with the same upper limit of $K_{\text{max}}=50$ bins and searches for the MDL-optimal bins.  
Fig.~\ref{fig:binning_strat} displays box-plots for the proposed \emph{quantile} method, the \emph{two-cuts}, and \emph{mid-points} heuristics of~\cite{kontkanen2007mdl}.  
We report
(i)~the bins count,  
(ii)~the \emph{MDL score}~\footnote{We refer to~\eqref{eq:mdl score} as the MDL score which quantifies the quality of the MDL histogram.},  
(iii)~the runtime (in minutes), and  
(iv)~the negative log-likelihood (NLL).
The MDL-score and NLL box-plots almost entirely overlap across the three strategies; medians and inter-quartile ranges coincide, indicating statistically similar fit quality. The runtime boxes, however, are spread over two orders of magnitude: the quantile method finishes in under 1 min, the mid-points grid centers around 20 min, and the two-cuts exceeds 70 min.
% The quantile method selects $20.4\!\pm\!1.1$ bins, achieves the lowest MDL score of $59\,390\!\pm\!92$, and finishes in only $0.74\!\pm\!0.09$ min.  
% By contrast, the two-cuts approach uses $19.0\!\pm\!1.2$ bins with a comparable NLL but requires $72.7\!\pm\!3.6$ min, while the midpoints strategy takes $20.4\!\pm\!0.9$ min.
These results indicate that restricting the candidate set via quantiles does not impose a significant loss in MDL optimality or likelihood fit, yet drastically improve computational efficiency.
% Thus, restricting the candidate set to empirical quantiles yields \emph{30–100\,$\times$} faster runtimes without sacrificing MDL optimality or likelihood fit.

Next, we evaluate our MDL-based PDF framework against a uniform discretization baseline following the set-up of~\cite{kargas2019learning}. Here, data are generated from five-dimensional Gaussian mixture ($N=5$, $R=6$), and the accuracy is measured by the KLD between the true and estimated densities, averaged over 100 Monte-Carlo trials.  
We compare the performance of our algorithm to that of the classical Gaussian mixture model based on EM (EM GMM), coupled tensor factorization algorithm based on alternating optimization and a KLD loss criterion (CTF-AO-KL)~\cite{kargas2019learning}, and the ``Oracle" method, which assumes that the labels (latent states) are known and serves as an empirical lower bound for the KLD. As shown in Fig.~\ref{fig:comparison}~(a),(b), our approach outperforms the uniformly binned (20 bins) CTF-AO-KL in both KLD and clustering accuracy. Fig.~\ref{fig:comparison}~(c) demonstrates how the histogram grows with $T$. When $T$ is small, adding many cuts would leave only a few points per bin, so the NML term cannot compensate for the MDL penalty. As more data become available, each prospective bin contains enough points to estimate its probability reliably; the NML term now rewards a finer partition, making
higher-resolution histogram both feasible and favorable.
% This results underscore the effectiveness of MDL-based binning in adapting to skewed and non-uniform densities, whereas uniform binning--due to its fixed allocation--~often leads to a model mismatch when the data deviate from uniformity. 
Although the EM GMM remains a practical choice for modeling Gaussian mixtures, its performance can be compromised if the discretization is not well adapted to the underlying density.

% \subsection{Real Data}
% \label{subsetc: real data}
\subsection{Real Data}
\label{subsetc: real data}

% \begin{table}[t]
%   \centering
%   \caption{\revOne{Performance of MDL-based binning strategies (mean ± std).}}
%   \label{tab:mdl-results}
%   \begin{tabular}{lccccc}
%     \toprule
%     Method  & \#Bins  & MDL score & Runtime [min] & NLL 
%     \\ 
%     \midrule
%     Proposed & $20.4\!\pm\!1.1$ & $59\,390\!\pm\!93$ & $\mathbf{0.74\!\pm\!0.09}$ & $13\,142\!\pm\!93$ 
%     \\
%     2 cuts & $19.0\!\pm\!1.2$ & $59\,423\!\pm\!92$ & $72.69\!\pm\!3.64$ & $13\,139\!\pm\!92$ 
%     \\
%     Midpoints& $19.3\!\pm\!1.1$ & $59\,417\!\pm\!92$ & $20.45\!\pm\!0.88$ & $13\,141\!\pm\!93$ 
%     \\ 
%     \bottomrule
%   \end{tabular}
% \end{table}

We further evaluate our proposed approach on a real dry bean dataset~\cite{koklu2020multiclass}.  In this experiment, each sample is characterized by features from 7 different dry bean varieties, and the objective is to accurately predict the bean class. Initially, we estimate the model rank using variational Bayesian inference (VB-PMF) \cite{chege2023bayesian}; given an upper bound for the rank $R$ (e.g. the maximum rank for which the Kruskal identifiability condition for the CPD is satisfied \cite{sidiropoulos2017tensor}), this algorithm automatically prunes irrelevant components to determine the effective rank. The estimated rank ($R = 48$) is then used to train our method and other competing algorithms.
We randomly split the data into training set ($80\,\%$) and testing set ($20\,\%$). We calculate the classification accuracy (defined as the proportion of correctly classified samples to the total number of samples) and report the results averaged over 50 random data splits. 
In general, as demonstrated in Table~\ref{tab:model_comparison}, our method achieves higher classification accuracy while also significantly reducing the computational cost.

\section{Conclusion}
\label{sect5}
In this paper, we introduced a unified framework that combines MDL and tensor factorization for non-parametric PDF estimation. By integrating a quantile-based cutting strategy, we improved the MDL computation time without compromising the performance. Experimental results on both synthetic and real datasets demonstrated the advantage of our approach compared to conventional uniform binning methods.

% table 2: Comparison of Models Based on Classification Accuracy
% \setlength{\tabcolsep}{5pt}
\begin{table}[t]
    \caption{Models Performance in Multiclass classification of dry beans (mean ± std).}
    \centering
    \label{tab:model_comparison}
    \begin{tabular}{lcc}
        \toprule
        Model & Class. Acc. & Runtime [min] \\
        \midrule
        VB-PMF & 86.85$\pm$0.62 & 8.67$\pm$0.56
        \\
        Proposed  & \textbf{87.72$\pm$0.67}  & \textbf{2.49$\pm$1.24}
        \\
        CTF-AO-KL & 87.04$\pm$0.59 & 75.3$\pm$10.9
        \\
        \bottomrule
    \end{tabular}
\end{table}
\bibliographystyle{IEEEtran}
% \clearpage
\bibliography{refs.bib}

\end{document}